# Surrogate Locally-Interpretable Models with Supervised Machine Learning Algorithms

Linwei Hu[1], Jie Chen, Vijayan N. Nair, and Agus Sudjianto

Corporate Model Risk, Wells Fargo, USA

July 9, 2020

## Abstract

Supervised Machine Learning (SML) algorithms, such as Gradient Boosting, Random Forest, and Neural Networks, have become popular in recent years due to their superior predictive performance over traditional statistical methods. However, their complexity makes the results hard to interpret without additional tools. There has been a lot of recent work in developing global and local diagnostics for interpreting SML models. In this paper, we propose a locally-interpretable model that takes the fitted ML response surface, partitions the predictor space using model-based regression trees, and fits interpretable main-effects models at each of the nodes. We adapt the algorithm to be efficient in dealing with high-dimensional predictors. While the main focus is on interpretability, the resulting surrogate model also has reasonably good predictive performance.

## 1 Introduction

Certain classes of supervised machine learning algorithms have much better predictive performance over traditional statistical methods. These include ensemble algorithms such as Gradient Boosting Machines (GBMs) and Random Forests (RFs) as well as neural networks (NNs). However, these models are complex and hence difficult to interpret without additional tools. In banking and finance, which are highly regulated environments, model interpretation is a critical requirement: users have to understand the model behavior, ensure consistency with business knowledge, and explain the results to customers, regulators, and other stakeholders. In response to these and related needs, researchers have been focusing on diagnostics for interpretability.

Some global diagnostics, such as variable importance analysis and partial dependence plots, have been around for a while (Friedman 2001). More recent developments include Sobol indices for global sensitivity analysis (Kucherenko, Tarantola, & Annoni, 2012), Global Shapely effects (Song, Nelson, & Staum, 2016), derivative based sensitivity (Kucherenko & others, 2010), ANOVA decomposition based on ICE plots (Chen et al. 2018). There are also various local importance tools, including LIME (Ribeiro, Singh, & Guestrin, 2016), Leave One Covariate Out (Lei, G'Sell, Rinaldo, Tibshirani, & Wasserman, 2018), SHAP explanation (Lundberg & Lee, 2017; Lundberg, Erion, & Lee, 2018), and Quantitative input influence (Datta, Sen, & Zick, 2016). Additional methods are available for neural network, including Integrated gradients (Sundararajan, Taly, & Yan, 2017), DeepLIFT (Shrikumar, Greenside, & Kundaje, 2017), and Layer-wise Relevance Propagation (Binder, Montavon, Lapuschkin, Müller, & Samek, 2016).

---

[1] Corresponding author (email: Linwei.Hu@wellsfargo.com)

This paper develops a new approach to interpretability called `Surrogate Locally-Interpretable Models' (SLIM). The idea is to treat the fitted values from the machine learning algorithms as the responses, use a model-based tree algorithm to do supervised partitioning of the space of predictors, and fit interpretable models within each node. The model-based tree algorithm fits main-effects only models to the predictors: linear regression or splines/GAM to account for non-linearity. Interactions among variables are accommodated by through the splitting variables. We describe post-processing diagnostics to understand the tree structure and the behavior of the models at leaf nodes and get insights into the overall results. Our studies show that SLIM often has good predictive performance on its own so that it can serve as a surrogate model too. (This is not the case if we fitted a single model-based tree to the original responses.) It turns out the fitting the model to surrogate responses (fitted values of an ML algorithm) leads to more stable trees, as the effect of random noise is reduced. Of course, the performance of the fitted tree is only as good as that of the underlying ML algorithm, so using surrogate responses from simple models will not lead to good results.

Our work takes its inspirations from two papers in the literature: Local Interpretable Model-Agnostic Explanations or LIME (Ribeiro, Singh, & Guestrin, 2016) and `born-again-trees' (Breiman and Shang 1997). LIME develops a local model around any point in the data space and is extensively used in ML applications. But separate models have to be developed for each point of interest, and one cannot easily understand how the local models change across the predictor space. KLIME was developed (Hall et al., 2017) as a global mode, but it used unsupervised partitioning (K-means and related algorithms) to cluster the predictor space. SLIM uses supervised partitioning and has superior performance to KLIME. Born-again trees (Breiman and Shang 1997) is one of the early efforts at interpreting machine learning algorithms. They used piecewise constant regression models which leads to deep trees that less interpretable. Model-based trees, used in SLIM, leads to shorter trees with fewer nodes.

The concept of `model-based trees' has been around since 1990s. M5 (Quinlan 1992) is probably the earliest algorithm, followed by others such as LOTUS ( Chan and Loh 2004), and party (Hothorn, Hornik, Strobl and Zeileis 2015). As Quinlan (1992) notes, model-based trees "are generally much smaller than [piecewise constant] regression trees and have proven more accurate in the tasks investigated".

The concept of fitting surrogate models is also known in the literature. In the computer experiments area, they are called emulators (see, for example, Bastos and O'Hagan 2009), and are used to approximate computationally-intensive physics-based models. Within the ML literature, they have also been called model distillation ( Tan, et al. 2018; Hinton, Vinyals and Dean 2015) and model compression ( Bucilua, Caruana and Niculescu-Mizil 2006).

The rest of the paper is organized as follows. Section 2 describes our SLIM method, diagnostics, and our efficient computational implementation. Sections 3 and 4 analyzes the performances, results, and interpretability of SLIM through simulation studies and real data sets. The paper ends with a summary in Section 5.

## 2 SLIM Methodology

### 2.1 Implementation of Model-Based Tree

The methodology is easy to describe since it is just based on model-based tree partitioning algorithms. Let $\{y_i, x_i, , i = 1, \ldots n\}$ be the original set of observations for the supervised learning problem: $y_i$'s are the (continuous or binary) univariate responses and $\boldsymbol{x}_i = (x_{1i}, \ldots, x_{Pi})$ is $P-$dimensional predictor. Let $\{\hat{y}_i = \hat{f}(\boldsymbol{x}_i), i = 1, \ldots, n\}$ be the predictions from the SML algorithm (random forest, GBM or NN) fitted to the original dataset. The $\hat{y}_i$'s will be used as the responses in the SLIM algorithm; to make this explicit, we will use the notation $y_i^S$ instead of $\hat{y}_i$ in the rest of the paper. Note that when the original response is binary, we will use the continuous predicted response $y_i^S$ in SLIM.

In addition, we may consider only a subset of the original $P-$dimensional predictors in the SLIM algorithm, for example, the top predictors identified through variable-importance analysis. To keep notation simple, we will continue to use the notations $\boldsymbol{x}_i$ for the predictor and $P$ for its dimension. Finally, we will use the same training and test datasets from the original SML algorithm: training dataset of size $n_1$ and test dataset of size $n_2$.

We will consider the following classes of main-effects models in building trees: linear regression with original predictors, linear models with spline-based transformation of predictors, and generalized-additive model (GAM), all without interactions. In practice, the relationships will be nonlinear and one will have to use the latter two. Now, any lack-of-fit in the model will be because of interactions, so the tree splitting will focus on interactions.

The algorithm is given below:

1. Let the root-node consists of all the observations in the training dataset $\{y_i^S, \boldsymbol{x}_i,, i = 1, \ldots n_1\}$; fit a main-effects model to the data; and compute the loss function or performance metric (typically sum of squared error (SSE) or SSE penalized by the effective degrees of freedom through the use of GCV (see Friedman, Hastie and Tibshirani 2009);
2. For $p = 1, \ldots P$: consider the predictor $x_p$ and its possible partitioning values – typically a set of quantiles $\{x_{1p}^* < x_{2p}^* < \cdots < x_{qp}^*\}$ for continuous predictors and all possible binary partitions for categorical predictors; partition the dataset into two sub-groups (nodes); fit the main-effects model to the two nodes, compute the within node performance metrics and overall combined metric; search over all possible partitions and predictor variables and determine the best predictor and split;
3. Compare the total value of metric for the two nodes with that for the parent node and determine if there is sufficient improvement according to the given stopping rule (such as reduction in SSE, maximum depth, or minimum number of observations in the child node). If there is improvement, split the observations according to this predictor and create two child node; if not stop;
4. If you split in step 3, repeat steps 2 and 3 until there is no further improvement or a stopping rule is met.
5. Prune back the tree using an appropriate model-fit statistic such as $R^2$, improvement in $R^2$, improvement in SSE, etc. These can be pre-determined thresholds or used as tuning parameters.
6. Optional: Once the structure of the tree is finalized, if the fitted models at the leaf nodes involve many predictors, one can refit all of the model in the nodes with L1 penalty.

In our implementation, we considered fitting the main-effects model with and without penalty. For the choice of penalty, at the tree-growing stage, the use of L2 penalty is effective in taking care of correlated predictors since it is computationally fast. The L1 penalty can be used in the final stage as noted in Step 6 above. Also, in our experience, spline transformations of the predictors with a small number of basis-functions are adequate in addressing nonlinearities.

## 2.2 Efficient Numerical Algorithm

The key step in fitting SLIM, or more generally any tree-based algorithm, is finding the best partition. This involves exhaustively searching all combinations of split variables and candidate split points (such as quantiles), and fitting two child models for each candidate split. This can be very time consuming in our applications which have many predictors. With 100 variables and 50 split points for each variable, we have to search over 5,000 possible combinations and fitting 10,000 child models. The naïve approach, which splits the data into two subsets according to each candidate split and fits the child models for each subset, would be computationally expensive. To address this, we developed an efficient algorithm for fitting linear models, one that reduces computations significantly. We also used parallelization with multiple CPU cores to reduce execution time.

Our algorithm has roots in other tree-based algorithms such as XGBoost. When searching for the best split point within a given splitting variable, XGBoost uses efficient steps: greedy algorithm, approximations, and histogram method, where the latter two are based on bucketing and aggregating the needed statistics in each bin. We use a similar strategy with some tweaks. Specifically:

1. For numerical split variables, use a set of split points, e.g., quantiles and bucket the split variable into bins. For categorical split variables, each category forms its own bin.
2. For each bin $k = 1,2,\ldots,K$, aggregate the statistics for fitting a main-effects model. For linear or spline-based regression, these are $XTX_k = \sum_{x_i \in bin_k} \tilde{x}_i^T \tilde{x}_i$, $XTY_k = \sum_{x_i \in bin_k} \tilde{x}_i^T y_i^S$, $YTY_k = \sum_{x_i \in bin_k} (y_i^S)^2$ where $\tilde{x}_i$ is the $i$-th row of the design matrix (after possible spline transformations and includes the intercept term), and $y_i^S$ is the SML prediction. These are called *gram matrices*.
3. For each split:
   a. Find the bins belonging to the left and right node, and calculate the sums $CXTX_L = \sum_{k \in L} XTX_k$, $CXTY_L = \sum_{k \in L} XTY_k$, $CYTY_L = \sum_{k \in L} YTY_k$; similarly calculate $CXTX_R, CXTY_R, CYTY_R$.
   b. The child models are fitted simply by $\beta_L = CXTX_L^- CXTY_L$ and $\beta_R = CXTX_R^- CXTY_R$ [2]. For ridge regression, roughly speaking, the coefficients are obtained as $\beta_L = (CXTX_L + \lambda I)^- CXTY_L$, but some manipulations to the gram matrices are required, e.g., to scale the variables.
   c. The errors can be computed from the gram matrices: $SSE_L = CYTY_L - 2\beta_L^T CXTY_L + \beta_L^T CXTX_L \beta_L$, with similar expressions for $SSE_R$. The penalized loss functions can be easily computed based on *SSE* and the effective degrees of freedom (GCV).

---

[2] The general matrix inverse is solved using eigenvalue decomposition $A = U^T D U$. Such decomposition is very helpful in ridge regression since $(A + \lambda I)^{-1} = U^T (D + \lambda I)^{-1} U$, which is very easy to compute given $D$ is diagonal, and can be reused for different $\lambda$ values.

Fitting the models using the pre-aggregated gram matrices reduces the computational complexity from $O(K(np^2 + p^3))$ to $O(np^2 + Kp^3)$, where $np^2$ is the complexity of calculating the gram matrix and $p^3$ is the complexity of solving for the parameters. When $n \gg p$, the gram matrix computation is the dominant part, so we have reduced the computation complexity by a factor of $K$. The errors *SSEs* are computed using small gram matrices, which also saves computation. One can further apply parallelization to reduce the computation time. It is straightforward to use an existing parallelization library (for example, joblib in python) to run different models for the split nodes in parallel.

We have implemented SLIM in python. It supports any user-defined main-effects model (linear regression, spline transformations, or GAM), with/without L2 penalty. In addition, we have built the interpretation methods discussed below inside the class to visualize the tree structure, plot the variable effects, calculate variable importance, and split the contributions. These are illustrated in Sections 3 and 4.

## 2.3 Diagnostics for Interpretability

### 2.3.1 Assessing Effects in Leaf Nodes

The fitted local models at each node $k$ are of the form:

$$h_k(\boldsymbol{x}) = b_{0k} + \sum_{j=1}^{p} h_{jk}(x_j)$$

consisting of all the $p$ important variables at that node. The model is additive with no interactions, so the input-output relationship between $x_j$ and fitted response is just $h_{jk}(x_j)$. We can easily visualize this to interpret the relationship. There are additional advantages to having an additive model.

To assess variable importance at each leaf node, we use the l*eaf-node sample variance* of predictor $x_j$ at leaf node $k$ denoted as $v_{jk} = \widehat{var}(\{h_{jk}(x_j)\})$ where $x_j = \{x_{j1}, \ldots, x_{jn_k}\}$ and $n_k$ is the number of observations in the $k-$th node. We recommend this simpler diagnostic rather than permutation-based importance scores which can be computationally expensive.

### 2.3.2 Assessing Global Interaction Effects

We also need to understand the structure of the tree and get insights into the splits. We can do this by quantifying the differences corresponding to all the non-split variables in the additive models before and after split. Specifically, suppose there are $p$ predictors $x_1, \ldots, x_p$, and assume the first splitting variable is $x_1$. The split occurs because of an interaction with one or more of the other variables. We want to examine the change in the behavior of each of the other predictors and quantify the difference (and hence the interactions). There are different ways to measure these differences in predictions. For linear regression models, the simplest is to look at the fitted coefficients of the predictors before and after the split. But this does not generalize to more general scenarios such as spline-based or additive models, so we describe a different approach below.

For a parent node $P$ and its left and right child nodes $L$ and $R$, denote the fitted additive models at these nodes as

$$h^P(\boldsymbol{x}_i) = b_{0P} + \sum_{j=1}^{K_P} h_{jP}(x_{ij}), i \in P,$$

$$h_L(\boldsymbol{x}_i) = b_{0L} + \sum_{j=1}^{K_L} h_{jL}(x_{ij}), i \in L, \quad \text{and} \quad h_R(\boldsymbol{x}_i) = b_{0R} + \sum_{j=1}^{K_R} h_{jR}(x_{ij}), i \in R$$

respectively. Here $i \in P$ denotes the $i$−th observation and $K_P$ denotes the number of predictors in the parent node $P$, with similar notation for the left and right child nodes. Now, for each predictor $x_j$ and observation $i$, consider the difference in the predicted models before and after the split:

$$d_{ij} = h_{jP}(x_{ij}) - h_{jL}(x_{ij}) \text{ for } i \in L \text{ and } h_{jP}(x_{ij}) - h_{jR}(x_{ij}) \text{ for } i \in R.$$

We can use the sample variance of these observations denoted as $c_j = var(\{d_{ij}\}|i \in P)$ as the effect of the split on variable $x_j$. To compare the effects across predictors, we normalize by the total variance to get the proportion contribution as $p_j = \frac{c_j}{\Sigma_j c_j}$. These diagnostics are illustrated in simulation and real data examples in Sections 4 and 5.

## 3  Illustration: Simulated Data

### 3.1  Case 1: Additive Model

We simulated 100,000 observations with $y = F_1(\boldsymbol{x}) + \epsilon$, where $F_1(\boldsymbol{x})$ is the model used in Tan et al. (2018):

$$F_1(\boldsymbol{x}) = 3x_1 + x_2^3 - \pi^{x_3} + \exp(-2x_4^2) + \frac{1}{2 + |x_5|} + x_6 \log|x_6|$$
$$+ \sqrt{2|x_7|} + \max(0, x_7) + x_8^4 + 2\cos(\pi x_8). \quad (1)$$

Here, $\boldsymbol{x}$ is a 10-dimensional predictor, $x_9$ and $x_{10}$ are noise variables not in equation (1), all iid $U(-1, 1)$, and $\epsilon \sim N(0, 0.5^2)$ for the 100,000 simulations. The training and test datasets were obtained by random splits of about 2/3 and 1/3. We fitted and tuned an XGBoost algorithm and used the predictions as surrogate responses.

Now we ran the SLIM algorithm using the 10 variables with XGBoost predictions as the response. To account for the nonlinearities in equation (1), we used linear B-spline transformations of the predictors with 15 quantile knots for each continuous variable. The model-based tree was run with a depth 2. We pruned the tree using the simple criteria: high $R^2$ ($> 0.99$) and/or low reduction of SSE ($< 2\%$ of the reduction in root-node SSE). Note that $R^2$ is just the squared correlation between the responses and their predicted values. This pruning left us with just one node (root) which is appropriate since this is an additive model.

Table 1 shows results on model performance: fidelity and accuracy. Here, fidelity means how well the surrogate model reproduced the original XGBoost predictions and accuracy means how well SLIM and XGBoost predicted their original responses. The fidelity values show that SLIM, our surrogate model, is very close to the original SML model with $R^2$ of 99.8%. The accuracy values indicate that SLIM has slightly worse performance on training data but better performance on test data, suggesting that there is less overfitting with SLIM.

Table 1. Performance Comparisons of SLIM and XGBoost

|  |  | MSE | R2 |
|---|---|---|---|
| **SLIM-Fidelity** | Train | 0.0124 | 0.998 |
|  | Test | 0.0132 | 0.998 |
| **SLIM-Accuracy** | Train | 0.250 | 0.958 |
|  | Test | 0.251 | 0.957 |
| **ML-Accuracy** | Train | 0.239 | 0.960 |
|  | Test | 0.265 | 0.955 |

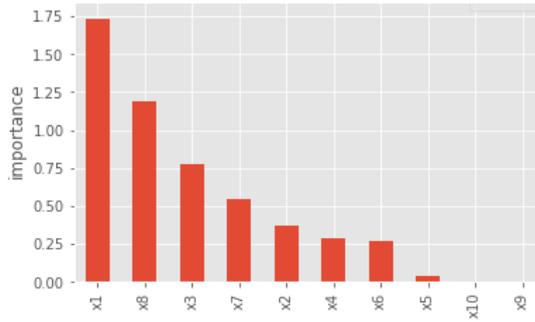

Figure 1. Variable importance (based on variances) for SLIM: Case 1

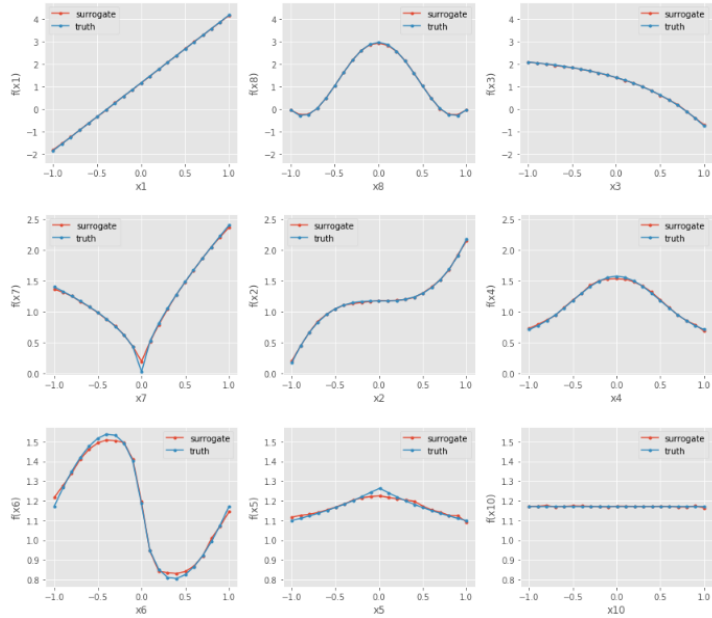

Figure 2. 1-D PDPs: SLIM vs true effects – Case 1

Since SLIM retains just the root-node, it just fits the model in equation (1) using spline-based transformations of predictors. For completeness, we briefly discuss the diagnostics on interpretability. Figure 1 is the variable importance plot; it shows that $x_1$ is the most important variable and the noise features $x_9, x_{10}$ are essentially 0. Figure 2 shows the 1-D

PDPs from SLIM overlaid with the true effects. The SLIM results are almost identical to the true model's variable effects and are consistent with results obtained in Tan et al (2018).

**3.2 Case 2: With Two-way and Three-way Interactions**

Next we added a few two-way interaction terms and a three-way interaction term to the model in equation (1) to get:

$$F_2(\boldsymbol{x}) = F_1(\boldsymbol{x}) + 2(I_{x_1>0})(I_{x_2>0})x_3 + 2(I_{x_1>0})x_4 + 4\big(x_5(I_{x_5>0})\big)^{|x_6|} + |x_7 + x_8|. \quad (2)$$

Again we simulated 100,000 observations with $y = F_2(\boldsymbol{x}) + \epsilon$, $\epsilon \sim N(0, 0.5^2)$, tuned an XGBoost algorithm and obtained the predictions. Then we fitted SLIM to these predictions as surrogate responses with tree depth = 5. As in Case 1, we fitted with linear B-spline transformations to the predictors. Table 2 shows the results from the full tree with depth = 5 (left) and the pruned tree (right). SLIM performs quite well on the full tree, and its test MSE is slightly better than XGBoost, suggesting that the latter may be overfitting slightly. However, this tree leads to 32 terminal nodes, and its challenging to interpret results from so many nodes. The pruned tree with only 12 nodes is shown in Figure 3. A few nodes on the fourth layer and all nodes on the fifth layer were pruned due to small values in *dsse* (reduction in SSE). These pruned nodes are shown in Figure 5. Pruning leads to some degradation in performance as seen in Table 2. It is a trade-off between performance and interpretability, one that users can manage by tuning the thresholds.

The display at each node in Figure 3 provides several pieces of information: number of observations (size), the splitting variable and split point, the *dsse* (reduction in SSE caused by the split), and $R^2$.

Table 2. Performance Comparisons of SLIM and XGBoost

|  |  | Full Tree with Tree Depth 5 | | Pruned Tree in Figure 3 | | |
| --- | --- | --- | --- | --- | --- | --- |
|  |  | MSE | R2 |  | MSE | R2 |
| **SLIM-Fidelity** | Train | 0.060 | 0.993 | Train | 0.115 | 0.987 |
|  | Test | 0.071 | 0.992 | Test | 0.123 | 0.985 |
| **SLIM-Accuracy** | Train | 0.283 | 0.968 | Train | 0.350 | 0.961 |
|  | Test | 0.289 | 0.967 | Test | 0.355 | 0.960 |
| **ML-Accuracy** | Train | 0.246 | 0.972 | Train | 0.246 | 0.972 |
|  | Test | 0.294 | 0.967 | Test | 0.294 | 0.967 |

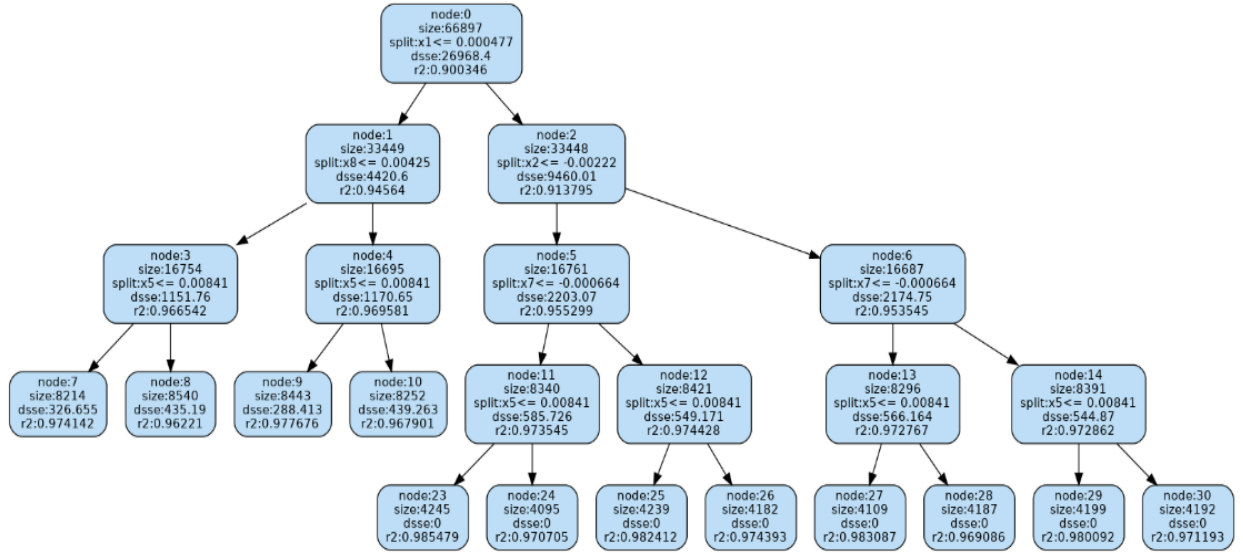

Figure 3. Tree structure of SLIM

To understand the structure of the tree, let us examine the partitions. The best split at the root-node N0 is based $x_1$ and the split occurs approximately at the point 0. This is consistent with the true model equation (2), and $x_1$ has an interactions with $x_2, x_3$, and $x_4$. In terms of the global interaction diagnostics in Section 2, the models for $x_4$ and $x_3$ had the largest changes after the split, with proportions of $p_4 = 80\%$ and $p_3 = 19\%$. All the others were less than 0.1%. These can also be seen by comparing the PDP plots of for $x_4$ and $x_3$ at N0, N1 and N2 in Figure 4.

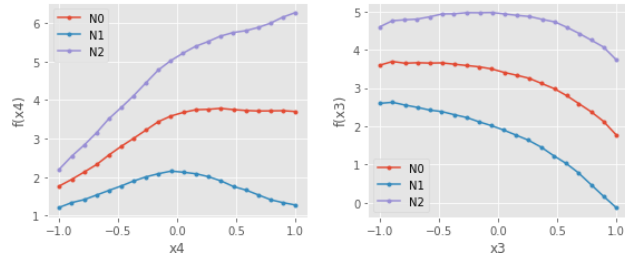

Figure 4. Comparison of 1D-PDP plots for $x_4$ and $x_3$ at N0, N1 and N2

Now N1 and its descendants correspond to observations with $x_1 \leq 0$. Since $I(x_1 > 0) = 0$ for these observations, the first two interaction terms in equation (2) are no longer active. N1 was further split on variable $x_8$, and the largest split contributor is $x_7$ with $p_7 > 99\%$ and the rest are all below 1%. So the driver for this split is the interaction between $x_8$ and $x_7$. Moving further down, both child nodes 3 and 4 are further partitioned based on $x_5$, again at (approximately) the value 0. This is also reasonable from equation (2). The pruned tree stops at this point. The unpruned tree (top left panel of Figure 5) shows that N7 is further split on $x_7$ and its child nodes 15 and 16 are again split, this time on $x_7$ and $x_8$ respectively, accounting for the $(x_7, x_8)$ interaction. Similarly, the top right panel shows that N8 is further partitioned on $x_5$ and the child nodes are both partitioned on $x_7$. These partitions account for the remaining parts of the global interactions in $(x_7, x_8)$ and $(x_5, x_6)$ that are not accounted for by the single splits in the pruned tree.

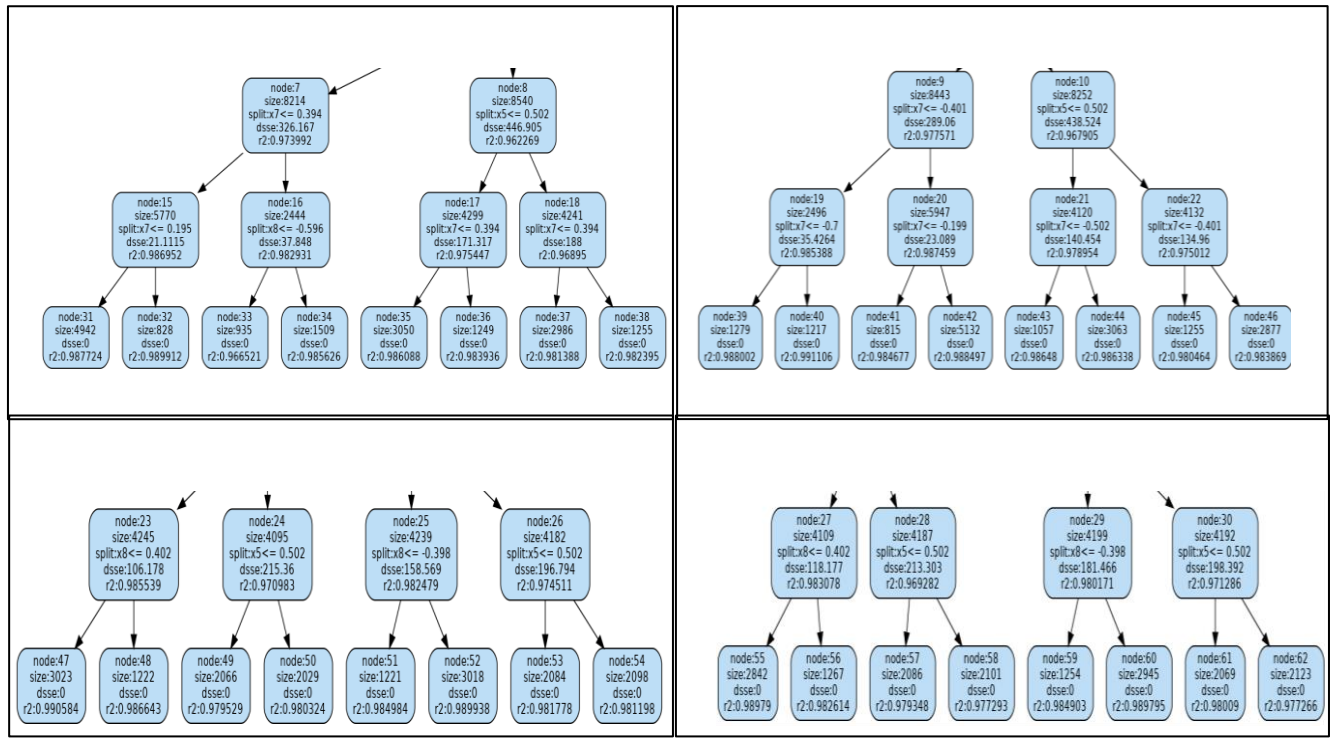

Figure 5. Nodes from the Full-Depth Tree that have been pruned: a) Top left: descendants of N7 and N8; b) Top right: descendants of N9 and N10; c) Bottom left: descendants of N23-N26; and d) ) Bottom right: descendants of N27-N30.

Next we follow N2 and its descendants where $I(x_1 > 0) = 1$. Now the effect of $x_4$ appears as a main effect; further, there is an interaction $I(x_2 > 0)x_3$. The next split is on variable $x_2$ which takes of this interaction. The interaction diagnostic also confirms this: the strongest driver is $x_3$ which accounts for almost 100% of the splitting effect. The next split is on variable $x_7$ to take care of the $(x_7, x_8)$ interaction followed by a split on $x_5$ to account for the $(x_4, x_5)$ interaction. The pruned tree stops at this point. The remaining (unpruned) nodes in the full tree are shown in the bottom left and right panels of Figure 5, and we see that the additional splits are based on variables $x_8$ and $x_5$ to account for the remaining parts of the global interactions. One can also plot the input-output relationships for the different nodes to examine the presence/absence of interactions. We do not include these diagnostics here because of space constraints.

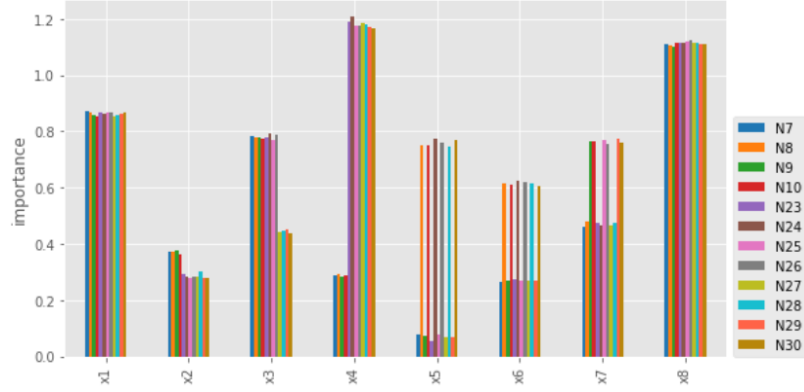

Figure 6. Variable importance (based on variances) at each leaf node for the eight active predictors

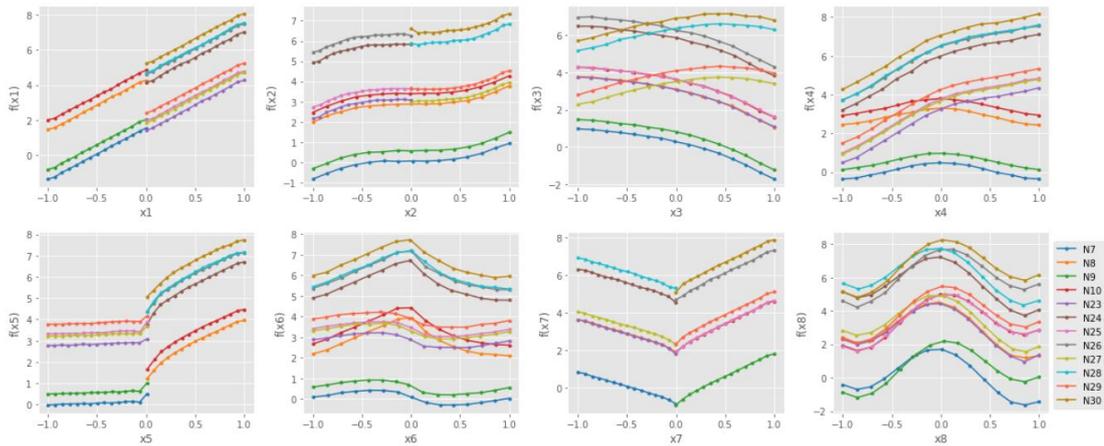

Figure 7. Variable effects at each leaf node for the eight active predictors

Figure 6 shows variable importance values for the 8 active variables (in equation 2) at the leaf nodes for the tree in Figure 3. The y-axis displays the variable importance measures, x-axis the predictors, and the color coding on the side indicates the nodes. Note that the importance of predictors variable $x_5, x_6$ and $x_7$ vary a lot among the terminal nodes. In particular, $x_5$ has low importance at nodes 7, 9, 23, 25, 27, and 29. Figure 7 shows the input-output relationships of these variables. We can see that the nonlinearities are adequately captured by B-spline transformations and interactions are also captured; for example the effects of $x_3$ are parallel across some nodes but not others. Note also that, for the top left for variable $x_1$, there are fewer curves for $x_1 < 0$ compared to $x_1 \geq 0$ since it appears as a splitting variable. The same phenomenon occurs for other splitting variables.

The takeaway point is that, despite the complexity of the global model in equation (2), the local models are very interpretable and the input-output relationships are easily explained. Further, the tree splits and associated diagnostics explain the nature of the interactions clearly.

## 4   Real Applications

### 4.1   Bikeshare data

This is a public dataset hosted on UCI machine learning repository that has been analyzed by Tan et al. (2018). The data has around 17,000 observations on hourly (and daily) bike rental counts along with weather and time information for 2011-12 in the Capital Bikeshare system.

Our goal is to predict hourly rental counts. We used log-counts as the responses in the model. Out of the original 17 predictors, we removed some non-meaningful ones, leaving us with 11: *mnth* (month = 1 to 12); *hr* (hour = 0 to 23); *holiday* (1 if yes and 0 otherwise); *weekday* (0 = sunday to 6 = saturday); *workingday* (1 if working and 0 if weekend or holiday); *season* (1:winter; 2:spring, 3:summer, 4:fall); *weathersit* (1:clear, 2: misty+cloudy; 3: light snow; 4 :heavy rain); *temp* (normalized with max of 41 C); *atemp* (ambient temp, normalized); *hum* (humidity); and *windspeed*.

We partitioned the dataset into a training set (2/3) and testing set (1/3), and used cross-validation to tune an XGBoost model on the training set. The variable importance scores (based on permutation) for the XGBoost model are given in Figure 8. Hour, working day, and temperature are the top three predictors.

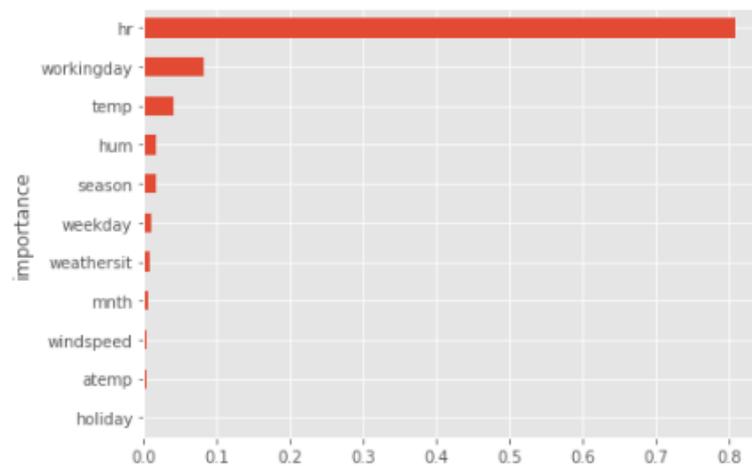

**Figure 8. Permutation-based variable importance scores for XGBoost**

The 1-D PDP for the top nine variables are shown in Figure 9. Note that working day has a tiny main effect (the range of variation in the y-axis is very small), which seems to contradict the variable importance result in Figure 8. The reason is that variable importance measures the overall effect (main effect and interaction); as it will become clear later, working day has a significant interaction with hour.

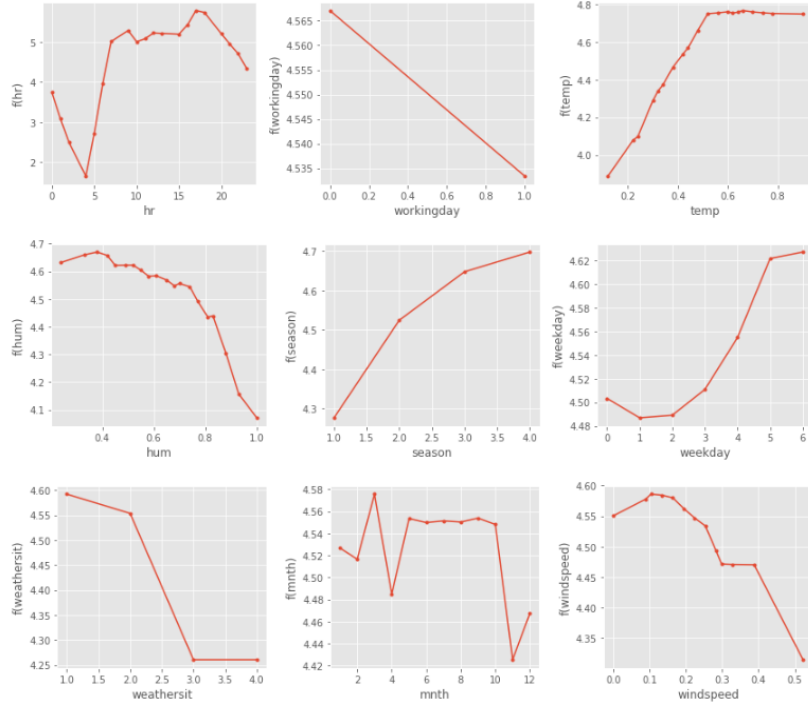

Figure 9. 1D-PDPs for top 9 variables in XGBoost model

The 2-D PDPs for the top four interaction pairs (identified through H statistics not shown here) are displayed in Figure 10. The largest interaction is between hour and working day (yes/no). For working day, the peak bike rentals happen around 7- am and 5-6pm whereas for non-working day, most rentals happen during 10am to 4pm. In addition, for both temp-hour and hum-hour interactions, the difference between low vs high temp/hum is largest during afternoon peak hour.

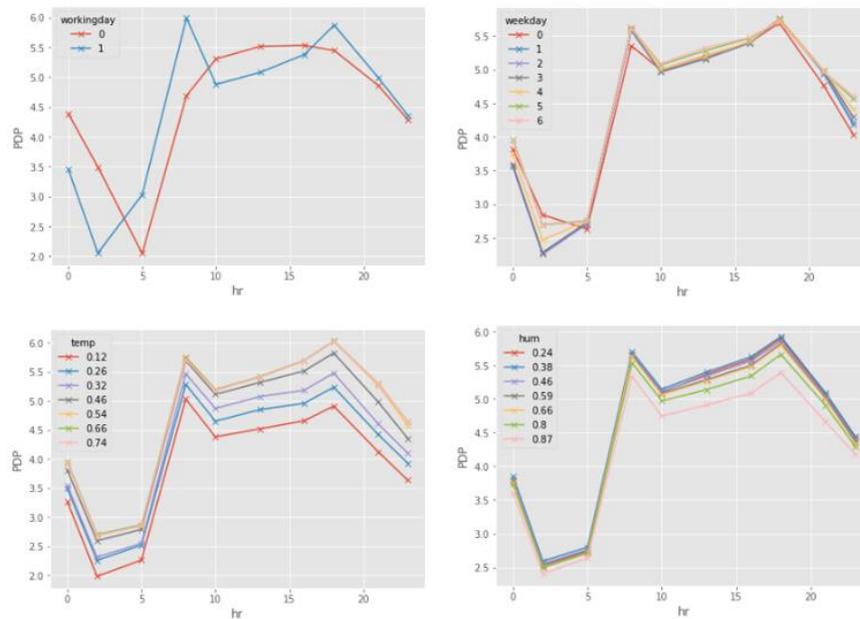

Figure 10. 2d PDPs for the top 4 interaction pairs in XGBoost model

Next we fitted the SLIM algorithm to the XGBoost predictions. We applied linear B-spline transformation with 25 quantile knots to each continuous variable and fitted SLIM with maximum depth of 3. The tree was pruned using $R^2$ and reduction in *sse*, as in Section 3.1. All nodes in the third layer and some nodes in the second layer were pruned, and the resulting tree is shown in Figure 11.

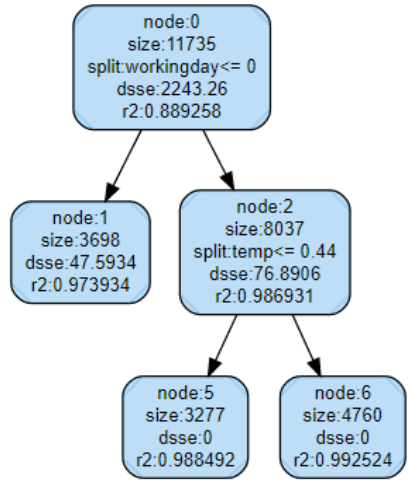

**Figure 11. Tree structure of SLIM**

The explanations from the SLIM tree in Figure 11 is relatively simple. The root-node N0 is split based on working day or not; N1 (right) to weekends/holidays and N2 (left) to working day. The largest driver for this split is its interaction with *hour* for which $p_j > 98\%$, which make sense. From the top left panel of Figure 13, we see that for non-working days, most rentals happen during 10 am – 4 pm. The pruned tree has no further split on N1, suggesting that an additive model is adequate for the remaining variables.

For N2, there is a further split based on temperature with the split at value of 0.44 (corresponding to about 16 degrees C). The largest drivers (and interactions) for the split are hr ($p_j = 53\%$), season ($p_j = 20\%$) and mnth ($p_j = 13\%$). Figure 13 shows that when temperature is low, there are fewer bike rentals. However, this difference is smaller during morning peak hour 7-9 am than at other times for working days. This is in fact a three-way interaction between working day, temperature and hour, and SLIM is able to capture that.

Table 3. Performances and Comparisons of SLIM vs XGBoost Algorithms

|  |  | MSE | R2 |
|---|---|---|---|
| **SLIM-Fidelity** | Train | 0.027 | 0.987 |
|  | Test | 0.028 | 0.986 |
| **SLIM-Accuracy** | Train | 0.187 | 0.916 |
|  | Test | 0.186 | 0.914 |
| **ML-Accuracy** | Train | 0.140 | 0.937 |
|  | Test | 0.157 | 0.928 |

Table 3 shows the performance results and comparisons with XGBoost. We have very good fidelity with $R^2$ about 98%. The accuracy is just slightly worse than the XGBoost model.

We can interpret SLIM model locally at each leaf node. Figure shows the variable importance at each leaf node. *Workingday* (= 0 or 1) is used in splitting, hence this variable becomes constant in each terminal node and has zero importance in leaf nodes. Similarly, holiday is constant for N5 and N6 since they correspond to *workingday* = 1. In all three leaf nodes, *hr* is the most important variable.

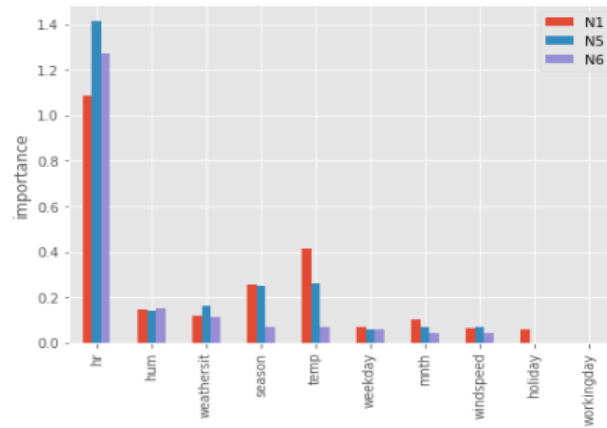

Figure 12. Variable importance (variance-based) at each leaf node

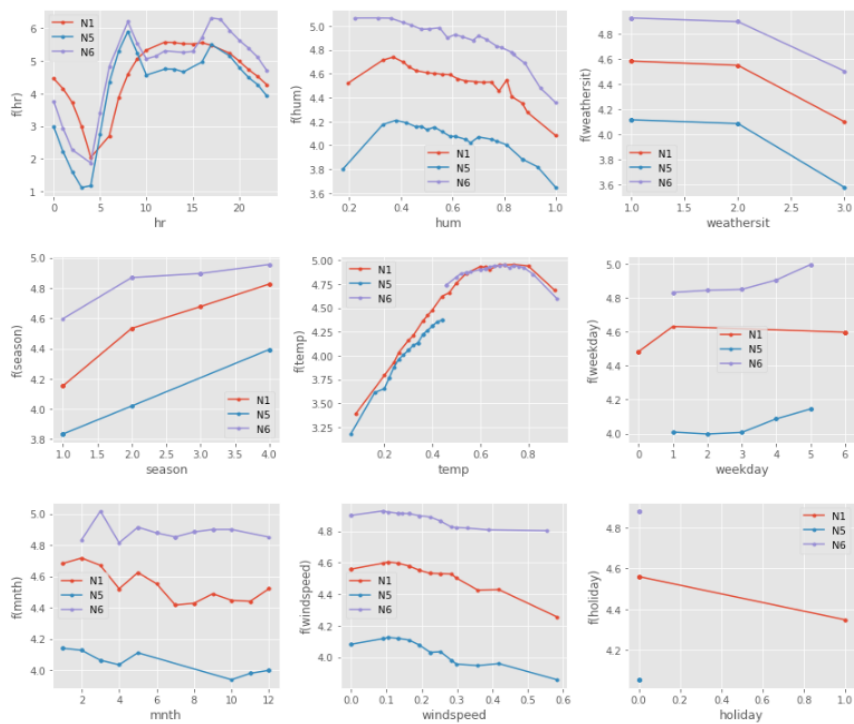

Figure 13. Variable effects at each leaf node

Figure 13 shows the input-output relationships in the additive models fitted at the leaf nodes. The nonlinearities are captured by B-spline transformations. In most panels, the curves are almost parallel, indicating that the effects are the same and are just offset by some constants. But in some panels, the shapes of the curves are different for N1 compared to N5 and N6. The most notable one is the top left panel where the curves for *hr* are different and we have already discussed this difference earlier. The less obvious one is the left panel in the middle row where the curves for season are also different. Note the curves for weekday seem to be different for N1 and N5, but they have little overlap (N1 is supported on 1-5 and N5 is support mostly on 0 and 6) so they are not considered as interaction.

We can compare the 1 and 2D-PDP plots for XGBoost in Figures 9 and 10 with the final fitted SLIM models in Figure 13 to assess the similarities and differences in interpretation. The *weekday* behavior in Figure 9 is somewhat different from that in Figure 13. There are small differences for the other variables as well. These are probably caused by aggregating disparate behaviors that are evident in the different nodes. The 2D-PDP plot in the top left panel of Figure 10 shows the differences in rental behavior between *workingday-by-hour* and is qualitatively similar to the average of the blue and purple curves in the top left panel of Figure 13. The latter provides more information by splitting the *workingday* behavior by temperature.

## 4.2 Home Mortgage Application

**Table 4. Variable definition for home lending data**

| Variable | Definition |
| --- | --- |
| fico0 (FICO) | FICO at prediction time |
| ltv_fcast (LTV) | loan to value (ltv) ratio forecasted |
| dlq_new_delq0 (del-status) | delinquency status: 1 means current 0 means delinquent |
| unemprt | unemployment rate |
| grossbal0 | gross loan balance |
| h | prediction horizon |
| premod_ind | time indicator: before 2007Q2 (financial crisis) vs after |
| qspread | spread, note rate - mortgage rate |
| sato2 | Spread at Time of Origination |
| totpersincyy | Total personal income year over year growth |
| orig_fico | origination fico |
| rgdpqq | real GDP year over year growth |
| orig_ltv | origination loan to value ratio |
| qspread2 | same as spread with minor difference |
| orig_cltv | origination combined ltv |
| orig_hpi | origination hpi (house price index) |
| balloon_in | balloon loan indicator |
| homesalesyy | home sales year over year growth |
| hpi0 | hpi at snapshot |
| homesalesqq | home sales quarter over quarter growth |

This example deals with home lending for residential mortgage. We used one million accounts (a subset) from a particular portfolio of mortgages. (The data have been modified for confidentiality reasons.) The response variable is binary indicating the loan defaulted or not. There were 50 predictors, and the key ones are listed in Table 4. We partitioned the data into training, validation and testing, each with 1/3 observations, trained an XGBoost model on the training set, and tuned the hyper parameters on the validation set. The performance was assessed on the testing set.

The variable importance scores for the top 20 variables of the XGBoost model are given in Figure 14. FICO, LTV and del-status are among the top three variables, and these are known to be important from the business context.

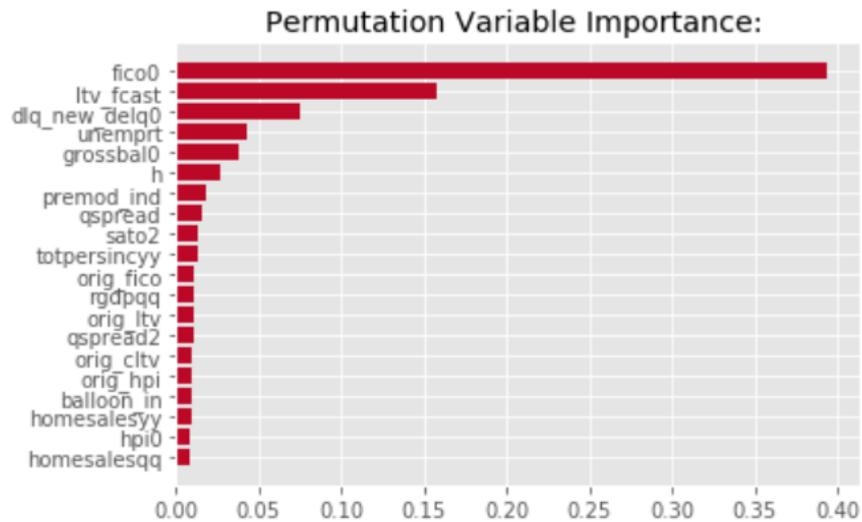

Figure 14. Permutation-based variable importance scores for XGBoost

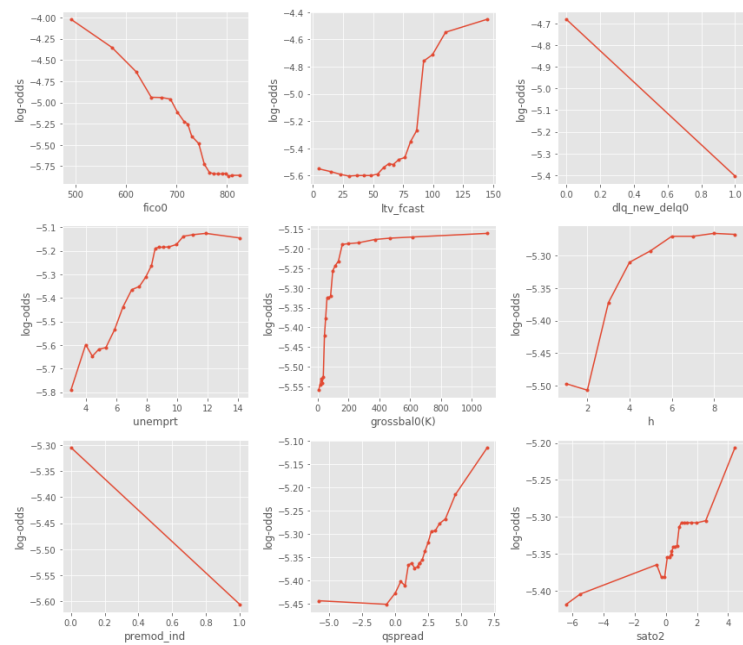

Figure 15. 1D-PDP for top 9 variables in XGBoost model

The 1-D PDPs for the top 9 variables are shown in Figure 15: y-axis shows the log-odds from the fitted model. As expected, default rate decreases as FICO and del-status increase, whereas LTV, horizon, unemprt, and grossbal have the opposite relationships. Most of the relationships are nonlinear. The one for FICO can be perhaps well approximated by a linear relationship except for very high FICO. Defaults increase sharply for LTV around 80-120, for *grossbal* near zero. There seems to be piecewise constant relationship for *qspread*, but this could be due to lack of data and poor fit in the lower tails.

The H-statistics for interactions among the top 9 variables are shown in 5. FICO, LTV and delinquency status have the strongest interaction effects.

Table 5. H-statistics among the top 9 variables in XGBoost model

|  | fico0 | ltv_fcast | dlq_new_delq0 | unemprt | grossbal0 | h | premod_ind | qspread | sato2 |
|---|---|---|---|---|---|---|---|---|---|
| fico0 | NaN | 0.10 | 0.18 | 0.04 | 0.03 | 0.01 | 0.12 | 0.01 | 0.01 |
| ltv_fcast | 0.10 | NaN | 0.26 | 0.02 | 0.00 | 0.01 | 0.01 | 0.02 | 0.01 |
| dlq_new_delq0 | 0.18 | 0.26 | NaN | 0.00 | 0.01 | 0.08 | 0.02 | 0.02 | 0.04 |
| unemprt | 0.04 | 0.02 | 0.00 | NaN | 0.01 | 0.02 | 0.00 | 0.01 | 0.01 |
| grossbal0 | 0.03 | 0.00 | 0.01 | 0.01 | NaN | 0.01 | 0.01 | 0.01 | 0.01 |
| h | 0.01 | 0.01 | 0.08 | 0.02 | 0.01 | NaN | 0.00 | 0.00 | 0.00 |
| premod_ind | 0.12 | 0.01 | 0.02 | 0.00 | 0.01 | 0.00 | NaN | 0.00 | 0.00 |
| qspread | 0.01 | 0.02 | 0.02 | 0.01 | 0.01 | 0.00 | 0.00 | NaN | 0.00 |
| sato2 | 0.01 | 0.01 | 0.04 | 0.01 | 0.01 | 0.00 | 0.00 | 0.00 | NaN |

The 2-D PDPs for the top 5 interaction pairs are shown in Figure 16: each curve in the plot shows the predicted log-odds against one variable conditional on the value of another variable, so non-parallel curves indicates interaction effects. Three out of the five plots show interactions with del-status. The effects of FICO and LTV are weaker when loan is in current status compared to delinquent status.

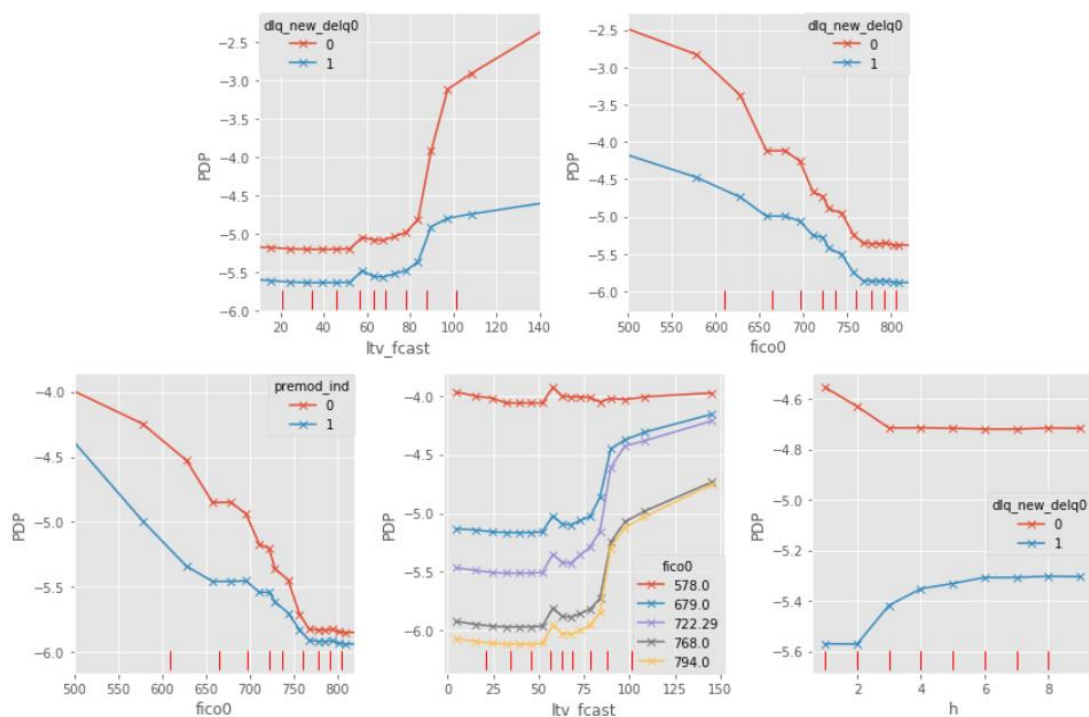

Figure 16. 2-D PDPs for top 5 interaction pairs

Now we fitted the SLIM algorithm to the logits of the XGBoost predicted probabilities. Only the top 20 variables in the XGBoost model, *qspread2* was removed due to extremely high correlation with qspread, leading to a total of 19 variables used in SLIM. We applied B-spline transformations with 20 quantile knots to each variable. Then we fitted SLIM with maximum depth 5. We pruned the tree using $R^2$ and reduction in *sse*, and the resulting tree structure is shown in Figure 17.

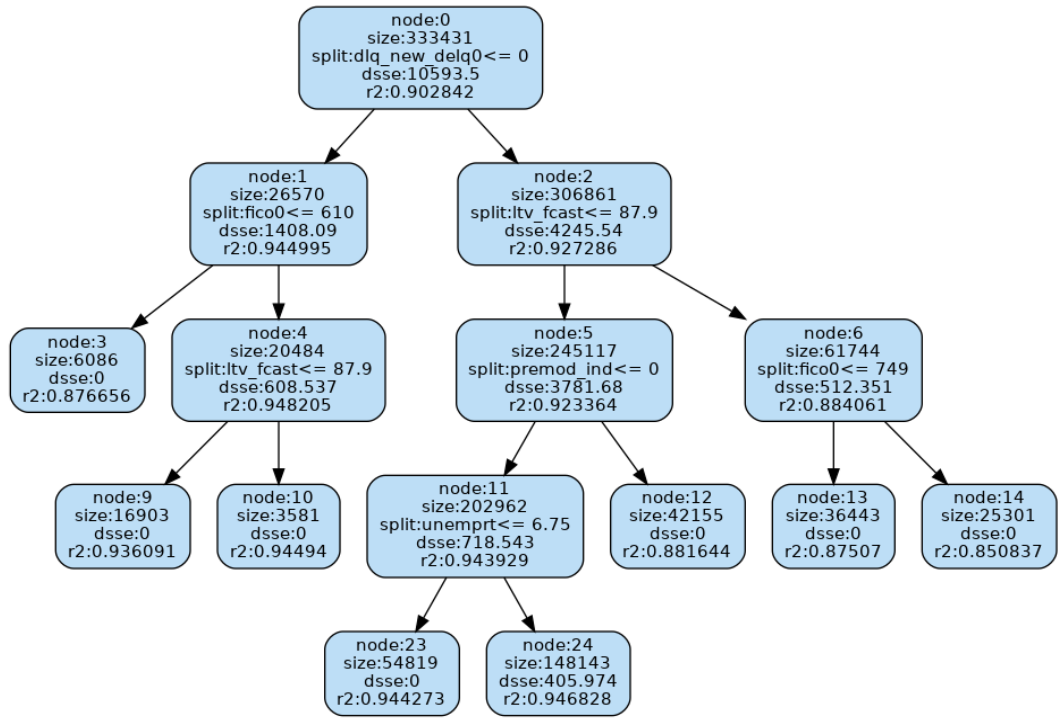

**Figure 17. Tree structure of SLIM**

The model performance is in listed in Table 6 and 7. We have very good fidelity, with $R^2$ about 97%. The accuracies (measured by both AUC and log-loss) are essentially the same as the XGBoost model.

**Table 6. Performance of SLIM and ML model**

|  |  | MSE | R2 |
|---|---|---|---|
| **SLIM-Fidelity** | Train | 0.032 | 0.968 |
|  | Test | 0.033 | 0.967 |

**Table 7. Performance of SLIM and ML model**

|  |  | AUC | log-loss |
|---|---|---|---|
| **SLIM-Accuracy** | Train | 0.816 | 0.043 |
|  | Test | 0.805 | 0.043 |
| **ML-Accuracy** | Train | 0.851 | 0.040 |
|  | Test | 0.805 | 0.043 |

Let us examine the tree structure in Figure 17. The splitting (interaction) variables are: {*def-status, FICO, LTV, pre-mod ind (before/after financial crisis), unemployment rate*}. The first split at N0 is based on del-status: accounts at N1 have del-status = delinquent and N2 have del-status = current, leading to separate models and more interpretable results. The three largest drivers for this split are FICO ($p_j = 48\%$), LTV ($p_j = 24\%$) and horizon ($p_j = 13\%$), and the rest are all negligible. This corresponds to three out of the five 2D interactions for XGBoost in Figure 16 and H statistics in Table 5. We can also conclude this from Figure 18 which compares the 1-D PDP plots of these three variables at N0 vs N1 and N2. The plot for N0 is very close to that for N2 because of the large number of observations in N2 compared to N1.

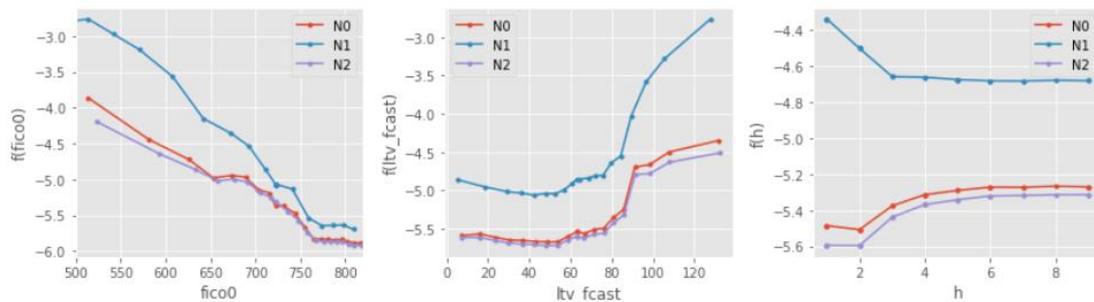

Figure 18. Comparison of variable effects for the split at N0

N1 is further split by FICO: N3 on the left has low (less than 640) and N4 to the right is high (greater than 640). The top two interactions drivers are LTV ($p_j = 54\%$) and h = prediction horizon ($p_j = 20\%$), and the rest are 10% or less. Note that FICO x LTV is one of the top interactions identified in Figure 16. The pruned tree does not split N3 further. N4 (higher FICO) is further split by LTV with N9 corresponding to counts with LTV lower than around 88 and N10 higher. The reduction in *SSE* is relatively small with this split. N2 (accounts that are current) is further split by LTV, and top two (interaction) drivers are FICO ($p_j = 44\%$) and balloon indicator ($p_j = 33\%$); the rest are below 10%. N5 and its child nodes go through further splits as well as N6, and one can examine the drivers for splitting in analogous manner. But the reductions in *SSE* are now much smaller.

Figure 19 shows the variable importance at each leaf node. Note that these scores are different for some predictors compared to the results in Figure 14 for XGBoost. Delinquency status is used in splitting at node N0, hence it is constant in the leaf nodes. FICO is still the most important predictor for all nodes except N14 which corresponds to high FICO (greater than 749). LTV is very important for N3 and N10 but less so for the other node. *Sato2* now appears as important for N3 and even more so N10, while it was relatively unimportant in XGBoost results.

The input-output relationships of the top nine variables, ranked by importance scores, are shown in Figure 20. The nonlinearities are captured by B-spline transformations. In addition, the effects of FICO, LTV and horizon are non-parallel across nodes, indicating different behavior across nodes. The others are mostly parallel, indicating more or less similar behavior for the remaining predictors across the leaf nodes.

SLIM leads to quite interpretable results at the leaf nodes. For example, N3 corresponds to delinquent accounts with low FICO. The most important predictors for explaining default probability in this node are (blue line in Figure 19): *LTV, prediction horizon (h), FICO, sato2, grossbal,* and *orig-hpi*. Figure 20 (blue curve) shows how default probability (log-odds) changes as the different predictors vary and compares them with their behavior at other nodes. For example, the top left panel in Figure 20 shows that default decreases as FICO increases with the decrease becoming sharper around 570, and stops at FICO = 610 which is the split point for N3. Similar interpretations can be made for the accounts at other nodes.

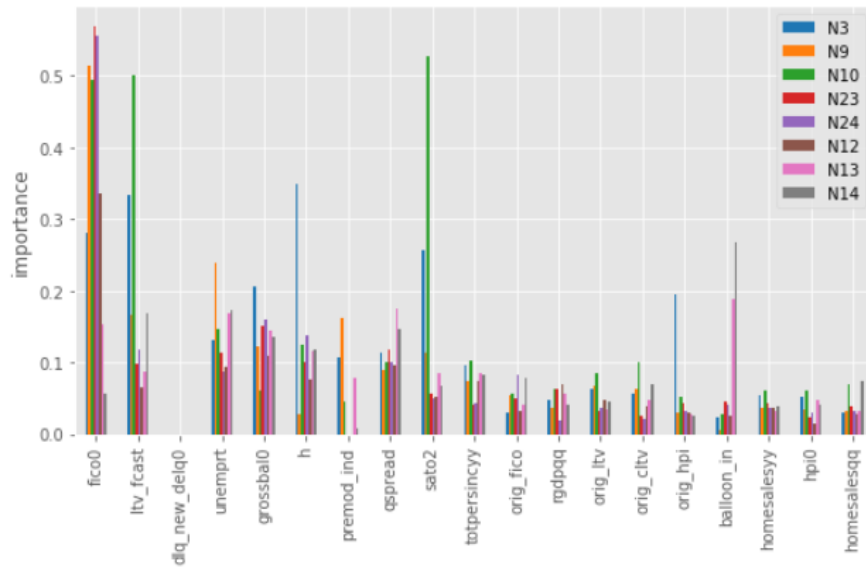

Figure 19. Variable importance (variance-based) at each leaf node

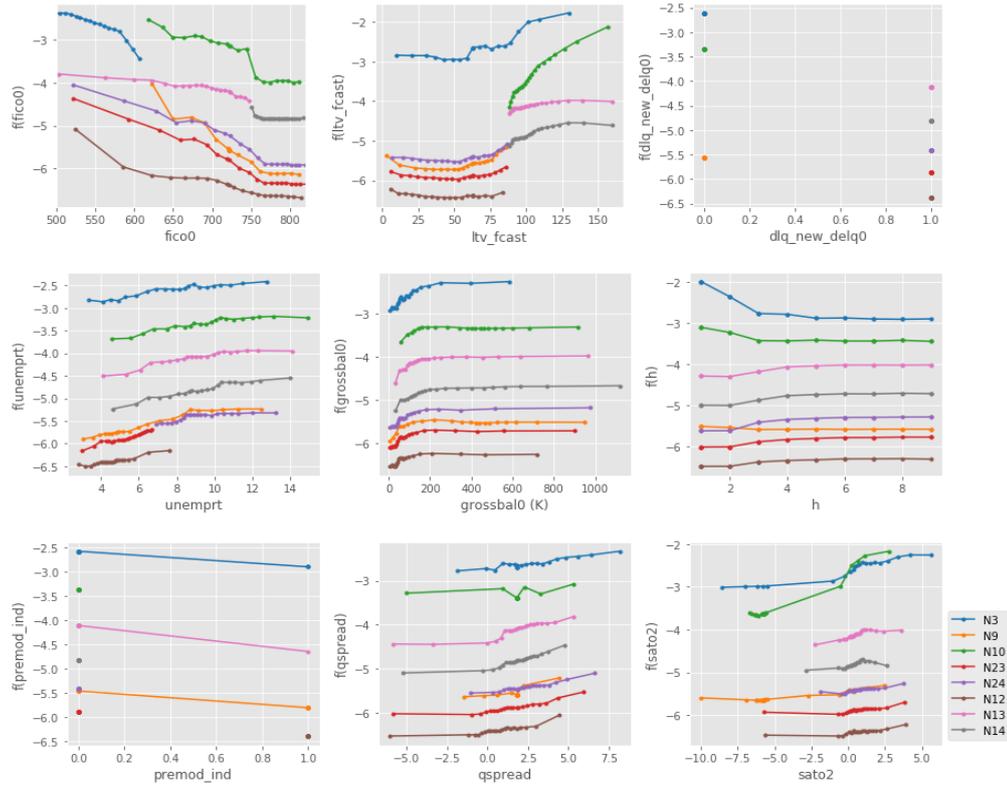

Figure 20. Variable effects at each leaf node

## 5 Concluding Remarks

We have introduced SLIM, an algorithm that fits a model-based tree on the responses of an ML algorithm to get locally interpretable results. The simulation and real data applications demonstrate the usefulness of the algorithm. SLIM is implemented using a computationally efficient algorithm and the associated diagnostics can be used to interpret the tree structure (drivers for the split) and also the final results. As noted earlier, trees fitted to the responses from the ML algorithm are more stable and they also have very good predictive performance, quite close to the original ML algorithm itself.

A limitation of SLIM, common to all tree-based regressions, is that the fitted model is not continuous at the boundaries, so predictions for points that are close to each other but in different partitions can vary considerably. We are currently working on using model-based trees to develop varying-coefficient models that can address this limitation.

**Acknowledgements**
The authors are grateful to Joel Vaughan and Xiaoyu Liu for their valuable comments during the course of this research.

The views expressed in the paper are those of the authors and do not represent the views of Wells Fargo.